\documentclass[letterpaper, 10 pt, journal]{IEEEtran}  

\usepackage{float}
\IEEEoverridecommandlockouts
\usepackage{cite}
\usepackage{amsmath,amssymb,amsfonts}
\usepackage{algorithmic}
\usepackage{graphicx}
\usepackage{textcomp}
\usepackage{xcolor}
\usepackage{todonotes}
\usepackage{bm}
\usepackage{comment}
\usepackage{tabularx,booktabs}
\newcolumntype{Y}{>{\centering\arraybackslash}X}
\usepackage{multirow}
\newcommand{\specialcell}[2][l]{%
  \begin{tabular}[#1]{@{}l@{}}#2\end{tabular}}

\newcommand{\mat}[1]{\mathbf{#1}}
\renewcommand{\vec}[1]{\boldsymbol{#1}}
\newcommand{\tI}{t}

\newcommand{\horiz}{T}
\newcommand{\state}{\mathbf{s}}       
\newcommand{\objectstate}{\state_o}
\newcommand{\robotstate}{\state_r}
\newcommand{\action}{\mathbf{a}}      
\newcommand{\solpolicy}{\hat{\policy}}             
\newcommand{\exppolicy}{\policy^*}            
\newcommand{\policyspace}{\Pi}             
\newcommand{\statedist}{d}             
\newcommand{\expectation}{\mathbb{E}} 
\newcommand{\truecost}{C}             
\newcommand{\on}{\sim} 
\newcommand{\costtoGo}{J}             
\newcommand{\loss}{\mathcal{L}}       
\newcommand{\affine}{\mat{T}}               
\newcommand{\taskspace}{\mathcal{T}}           
\newcommand{\se}[1]{\mathrm{SE}(#1)}      
\newcommand{\argmin}{\operatorname*{arg\,min}}
\newcommand{\argmax}{\operatorname*{arg\,max}}

\newcommand{\combined}{\hat{\policy}_{\objectstate}}
\newcommand{\filt}{\mathbf{F}}

\newcommand{\DI}{t}
\newcommand{\NI}{k}
\newcommand{\maxK}{K}
\newcommand{\infK}{K_{\text{inf}}}
\newcommand{\TIscale}{c}

\newcommand{\obsfeature}{\mathbf{z}_{o,\DI}}

\newcommand{\period}{T}

\newcommand{\obsperiod}{\period_o}

\newcommand{\actionpred}{\mathbf{A}^\NI_\DI}
\newcommand{\actionind}[1]{\mathbf{A}^{#1}_\DI}
\newcommand{\noise}{\epsilon}
\newcommand{\params}{\theta}
\newcommand{\noisenet}{\noise_\params}

\newcommand{\cnn}{\mathbf{E}}

\newcommand{\vfencoder}{\mathbf{VF}\cnn}

\newcommand{\proprio}{\vec{x}}
\newcommand{\shearfield}{\boldsymbol{\mathcal{U}}}

\newcommand{\policy}{\pi}

\newcommand{\horizontal}{\text{x}}
\newcommand{\vertical}{\text{y}}

\newcommand{\image}{\mat{I}}
\newcommand{\optflowfunc}{\mathbf{Flow}}

\newcommand{\ihp}{\mathbf{IHP}}
\newcommand{\surrloss}{\ell}

\usepackage{array}
\newcolumntype{P}[1]{>{\centering\arraybackslash}p{#1}}

\def\BibTeX{{\rm B\kern-.05em{\sc i\kern-.025em b}\kern-.08em
    T\kern-.1667em\lower.7ex\hbox{E}\kern-.125emX}}

\title{
GrOMP: Grasped Object Manifold Projection for Multimodal Imitation Learning of Manipulation
}

\author{William van den Bogert$^{1}$, Gregory Linkowski$^{2}$, and Nima Fazeli$^{3}$
\thanks{$^{1}$ William van den Bogert is with the Mechanical Engineering Department at the University of Michigan, MI, USA 
{\tt\small willvdb@umich.edu}}%
\thanks{$^{1}$ Gregory Linkowski is with the Robotics \& Automation Research Department at the Ford Motor Company, MI, USA 
{\tt\small glinkows@ford.com}}%
\thanks{$^{3}$ Nima Fazeli is with the Robotics Department at the University of Michigan, MI, USA
{\tt\small nfz@umich.edu}}%
}
\begin{document}


\maketitle
\thispagestyle{empty}
\pagestyle{empty}

\begin{abstract}
Imitation Learning (IL) holds great potential for learning repetitive manipulation tasks, such as those in industrial assembly. However, its effectiveness is often limited by insufficient trajectory precision due to compounding errors. In this paper, we introduce Grasped Object Manifold Projection (GrOMP), an interactive method that mitigates these errors by constraining a non-rigidly grasped object to a lower-dimensional manifold. GrOMP assumes a precise task in which a manipulator holds an object that may shift within the grasp in an observable manner and must be mated with a grounded part. Crucially, all GrOMP enhancements are learned from the same expert dataset used to train the base IL policy, and are adjusted with an $n$-arm bandit-based interactive component. We propose a theoretical basis for GrOMP's improvement upon the well-known compounding error bound in IL literature. We demonstrate the framework on four precise assembly tasks using tactile feedback, and note that the approach remains modality-agnostic. Data and videos are available at williamvdb.github.io/GrOMPsite.
\end{abstract}

\section{Introduction}
Imitation learning (IL) is a powerful tool for generating complex manipulator behavior for repeatable tasks. Diffusion-based behavior cloning has enabled IL to generate multi-modal trajectories, solving a pervasive interpolation problem \cite{chi2023}. However, all IL methods suffer from compounding errors \cite{block2024}: as the policy rolls out, small errors in the policy compound resulting in deviation from desired behavior. As such, there are no guarantees that learned high-precision assembly tasks can be as successful as the manufacturing industry requires.



Today, assembly automation is highly dependent on specialized fixtures and end effectors to ensure repeatable performance. These highly structured environments are extremely effective, but also incur a high financial cost if designs are upgraded and fixtures are changed to fit modified parts. The more flexible automation approach we consider in this paper involves IL for manipulation with a non-specialized gripper, such that we cannot assume a rigid grasp on an object where in-hand object poses also incur compounding errors.

\begin{figure}[t]
\centerline{\includegraphics[width=\columnwidth]{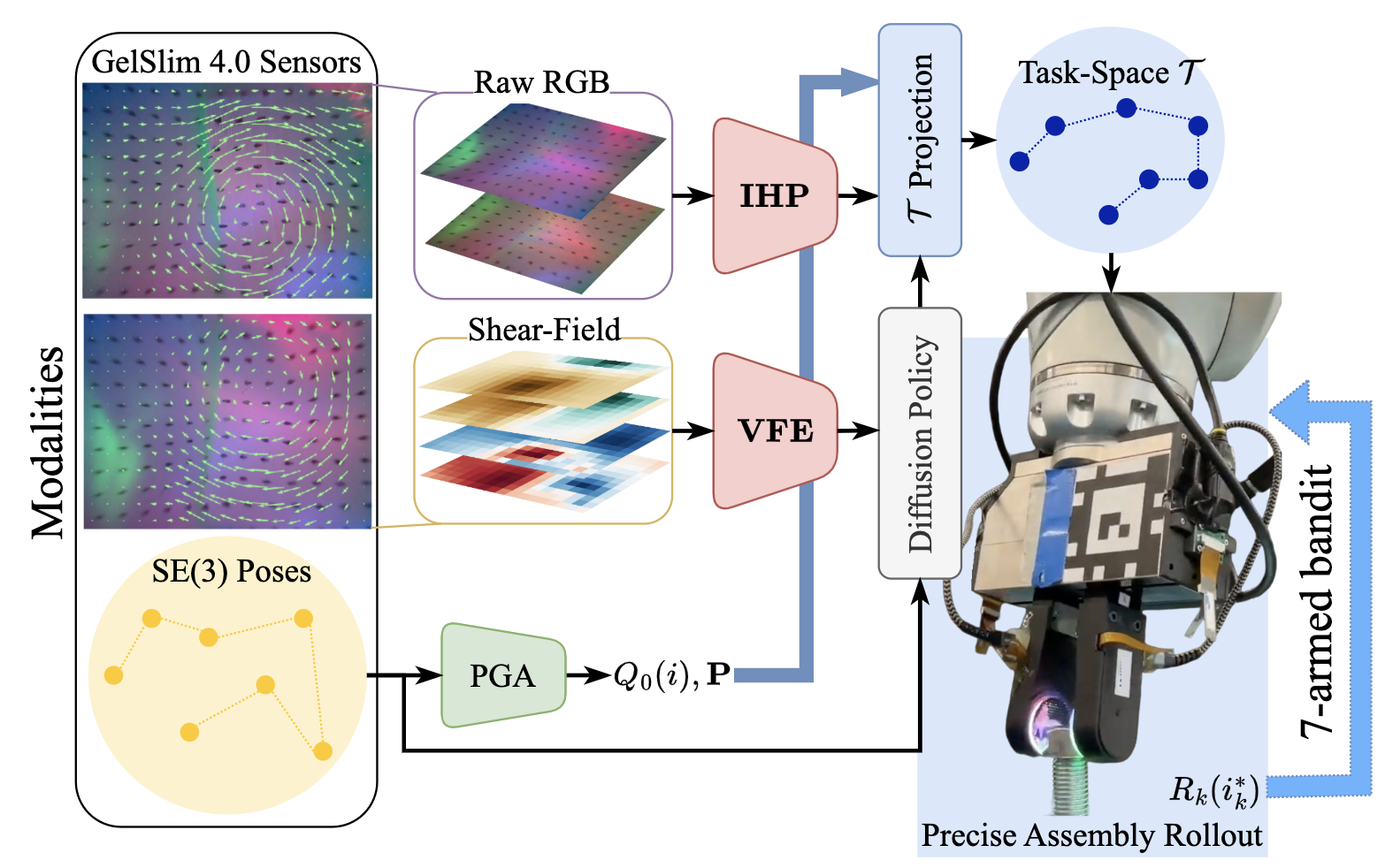}}
\caption{An overview of Grasped Object Manifold Projection as implemented in this paper. Vision-based tactile sensors (GelSlim 4.0) provide a field of shear displacements and raw RGB images. Shear-fields and proprioception serve as modalities for Diffusion Policy (DP), while RGB images are used for in-hand object pose estimation ($\ihp$). DP trajectories and object poses are used to project robot-driven grasped object behavior to the task space $\taskspace$, derived from principal geodesic analysis (PGA) of the expert dataset. A 7-armed bandit adjusts this projection based on rollout rewards.}
\label{fig:teaser}
\end{figure}

To address this challenge, we introduce \textbf{Grasped Object Manifold Projection (GrOMP)}, an interactive framework which operates on top of an IL policy to constrain a grasped object to a lower dimensional task space, as outlined in Fig \ref{fig:teaser}. Our method learns a task-space manifold from expert demonstrations and projects IL trajectories to this manifold, removing compounding errors orthogonal to its tangent space. We also provide an interactive reinforcement learning method for selecting the manifold based on observed successes. Finally, we demonstrate GrOMP against vanilla IL---implemented as a Diffusion Policy \cite{chi2023}---on real robot experiments.

\section{Related Works}
Perhaps the most influential interactive behavior cloning algorithms is DAgger \cite{ross2011_dagger}, which continuously augments the training dataset with the resulting policy rollouts, with the option to query the expert. DAgger demonstrated theoretically and in practice an improvement upon the bound in Eq. \ref{eq:costtogo}, and went on to inspire a host of variants. For instance, MEGA-DAgger \cite{sun2024} queries multiple imperfect experts. Human-Gated DAgger (HG-DAgger) \cite{kel2019} allows a human expert to invoke doubt to the novice before contributing to the dataset. These methods are tested on autonomous driving rather than manipulation. Diffusion-Meets-DAgger (DMD) \cite{zha2024} is well-tested on  eye-in-hand manipulation tasks and uses diffusion to produce novel task views that become part of the DAgger dataset. DAgger and its variants provide inspiration for this work.

GrOMP does not rely on dataset aggregation as DAgger does. Rather, it assumes an existing, constant, task-specific manipulation dataset, and applies extra understanding of a task manifold to the behavior-cloned policy. This is inspired by several Bayesian and reward-based methods. DropoutDAgger \cite{ken2017} ensures the predicted action is sufficiently close to the expert action and invokes the expert if not. SQIL \cite{red2019} provides simple rewards for novice behavior that is closer to the expert. CCIL \cite{ke2024_ccil} also maintains closeness to the expert by generating corrective labels for training. These methods serve a similar purpose to the projection onto the task manifold learned from expert demonstration as in GrOMP. GrOMP bases its adjustment of this manifold on classical RL, which is used in conjunction with IL in other methods. Inverse Reinforcement Learning (IRL) is a classic method which learns a reward function for RL from demonstration, while Generative Adversarial Imitation Learning (GAIL) \cite{ho2016} extracts a policy directly from demonstrations as if IRL was used.

The state of the art for ``vanilla" behavior cloning (imitation learning via learning the state-action map) is Diffusion Policy (DP) \cite{chi2023}, where the state-action map is a generative diffusion model. This method has been shown to require $>$100 demonstrations for manipulation tasks that require medium precision \cite{chi2023}. For GrOMP, we show an improvement upon DP for high precision assembly tasks. In this paper, we demonstrate this with tactile feedback. While DP was initially only tested using visual feedback, it lends itself well to multimodal feedback and has since been tested with tactile feedback, including vision-based tactile feedback. While \cite{geo2024} found DP with tactile feedback to produce zero successful USB insertions, we find some success, with greater success when GrOMP is used.

\section{Preliminaries}\label{preliminaries}

\subsection{Problem Statement}\label{prob}

We assume a manipulator robot is tasked with learning a precise assembly task, wherein one part grasped by its end effector must be mated to another which is fixed in the workspace. We consider a non-rigid grasp. That is to say the transformation $\affine_{to}(t) \in \se{3}$ between the end-effector and the object is not constant, thus we define separately the object state $\objectstate \supseteq \affine_{so}(t)$ and the robot state $\robotstate \supseteq \affine_{st}(t)$. In words, both states contain their respective poses, but there may be other elements i.e. forces, contacts, velocity. While the ground truth states are unknown, we assume access to observations (in this paper specifically, robot proprioception and high-resolution tactile sensing at the fingertips).


Under these assumptions, the manipulator must learn to perform this mating from a distribution of initial conditions throughout the demonstrations presented for supervised imitation learning (IL). As with all behavior cloning, we seek the resulting policy $\solpolicy$ as a state-action map, where states can consist of multi-modal observations, and actions consist of robot end-effector trajectories of $\affine_{st}(t) \in \se{3}$.

\subsection{Background and Motivation}\label{prob}

As presented in the seminal works of Ross et al. \cite{ross2010,ross2011_dagger}, an imitation learning policy can be derived using the following supervised learning formulation:
\begin{equation}
    \solpolicy = \argmin_{\policy \in \policyspace} \expectation_{\robotstate \on \statedist_{\exppolicy}}[\surrloss(\robotstate,\policy)]\label{eq:supervised}
\end{equation}
where $\policyspace$ is the policy class, $\statedist_{\exppolicy}=\sum_{\tI=1}^{\horiz} \statedist^\tI_{\policy}$ is the distribution of states under the expert policy $\exppolicy$ after $\horiz$ steps, and $\surrloss$ is the observed surrogate loss, which is minimized instead of $\truecost(\state,\action)$, the true cost for the particular task. Ross et. al \cite{ross2011_dagger} showed a bound for the cost-to-go $\costtoGo(\policy) = \sum_{\tI=1}^{\horiz} \expectation_{\robotstate \on \statedist^\tI_{\policy}}[\truecost_\policy(\robotstate)]$ for the policy $\policy$ where $\truecost_\policy(\robotstate) = \expectation_{\action \on \policy(\robotstate)}[\truecost(\robotstate,\action)]$ is the immediate cost of enacting the policy $\policy$ in the state $\robotstate$. Specifically, by assuming $\surrloss(\robotstate,\policy)$ is the expected 0-1 loss of $\policy$ with respect to $\exppolicy$, \cite{ross2010} showed the following bound: 


\begin{equation}
    \costtoGo(\policy) \leq \costtoGo(\exppolicy) + \horiz^2\noise\label{eq:costtogo}
\end{equation}
Where $\noise = \expectation_{\robotstate \on \statedist_{\exppolicy}}[\surrloss(\robotstate,\policy)]$. 
For our problem of learning precise assembly with a non-rigid grasp, we are most interested in controlling the object state $\objectstate$ and reformulate the expert policy relying on $\objectstate$:
\begin{equation}
    \exppolicy = \argmin_{\policy \in \policyspace} \expectation_{\objectstate \on \statedist_{\policy}}[\truecost_\policy(\objectstate)]\label{eq:expert}
\end{equation}
This reformulation potentially worsens the bound given the uncertainty in the robot commanded actions (our input) and the relative object motion in the grasp (our desired output), particularly due to contacts. 

This formulation implies that compounding errors accumulate across all states \cite{ross2010}. However, many tasks, particularly those encountered in manufacturing evolve over lower-dimensional manifolds in the state space. These manifolds are induced by the mechanical constraints of a task (e.g., screw motion for nut and bolt assembly or slide-to-insert for peg-insertion). Inspired by this insight, we propose \textbf{Grasped Object Manifold Projection (GrOMP)}, which constrains the object to a lower-dimensional task space manifold $\taskspace$ during IL rollout. Our method results in a combined policy $\combined$ that provides corrections to the base IL policy via projections onto this manifold. Given that the manifold $\taskspace$ is derived from observations of the task-specific expert policy $\exppolicy$, we assume $\taskspace$ is representative of the task completed by $\exppolicy$. Thus, we expect the compounding error represented by Eq. \ref{eq:costtogo} are mitigated along directions orthogonal to $\taskspace$.

\section{Method}\label{method}
In this section, we describe our instantiation of GrOMP, first describing the projection which is applied to the IL trajectories, then outlining the selection process of the projection dimensionality from the expert demonstrations, and the $n$-arm bandit method for adjusting the projection selection from the resulting success rates.

\subsection{Task Manifold Projection} \label{proj_filter}

Our goal is to derive a behavior cloned policy using the formulation from Ross et al. (Eq. \ref{eq:supervised}). GrOMP supplements this problem with the mapping $\filt_{\objectstate}:\se{3}\rightarrow\se{3}$ which is computed from a lower-dimensional task space manifold $\mathcal{T}  \subseteq \se{3}$ that is representative of the task. This policy seeks the robot pose $\affine_{st}$ such that the object pose is projected to the task space $\taskspace$ which is learned from the demonstration dataset using the method described in Section \ref{selection}. We formalize this computation as:
\begin{equation}\label{eq:taskspaceconstraint}
\affine_{st} = \filt_{\objectstate}(\solpolicy(\robotstate)) = \filt_{\objectstate}(\affine_{st}^{\solpolicy}) = (\text{proj}_\taskspace\affine_{so})\affine_{to}^{-1}
\end{equation}
In words, we first project the object pose w.r.t. the world-frame into the task manifold ($\text{proj}_\taskspace\affine_{so}$), then compute the goal robot pose from the estimated object pose w.r.t. the robot.
We assume that the task demonstrations are sufficiently informative to recover $\mathcal{T}$. To illustrate the importance of formulating the object pose (as opposed to robot pose) constraint on the task manifold $\mathcal{T}$, we imagine a task where the goal is to keep the object stationary w.r.t. the world frame. For such a task, $\text{dim}(\mathcal{T})=0$. However, the robot may move in a subset of $\se{3}$ that will allow the object to be stationary due to the possibility of relative slip.
In this case of $\text{dim}(\mathcal{T})=0$, $\affine_{st}=\affine_{to}^{-1}$ so that $\affine_{so}=\mat{I}$. 
We assume for our method that $\affine_{to}$ is measured via tactile sensors. Alongside the manifold constraint, the IL policy $\solpolicy$ would be responsible for manipulating the object within the task space to complete the task.

\subsection{Deriving Task Space from Demonstration}\label{selection}
While a task space could be manually selected, in this section we present a method whereby this manifold can be learned from demonstration using principal geodesic analysis (PGA) \cite{PGA}. To perform PGA, we first project the $\se{3}$ object poses in the expert demonstration dataset to the tangent space at the mean, using the logarithm map:
\begin{align}
&\vec{\xi}^* = \begin{bmatrix}
        \vec{\omega}\\
        \vec{v}
    \end{bmatrix}\text{ where }\hat{\vec{\xi}}=\log(\mat{(\overline{\affine_{so}^{\exppolicy}})^{-1}\affine_{so}^{\exppolicy}})=\begin{bmatrix}
        \vec{\hat{\omega}} & \vec{v}\\
        \mat{0}_{1\times3} & 0
    \end{bmatrix}
\label{eq:convert}
\end{align}
where these $\vec{\xi}^*\in\mathbb{R}^6$ are the twists and $\overline{\affine_{so}^{\exppolicy}}$ represents the geodesic mean (calculated as in \cite{PGA}) on the manifold of $\se3$ of all object poses in the expert dataset with respect to the world frame, and $\vec{\hat{\omega}}$ represents the skew-symmetric matrix representation of $\vec{\omega}$ \cite{math_robot}.

\begin{figure}[h]
\centerline{\includegraphics[width=\columnwidth]{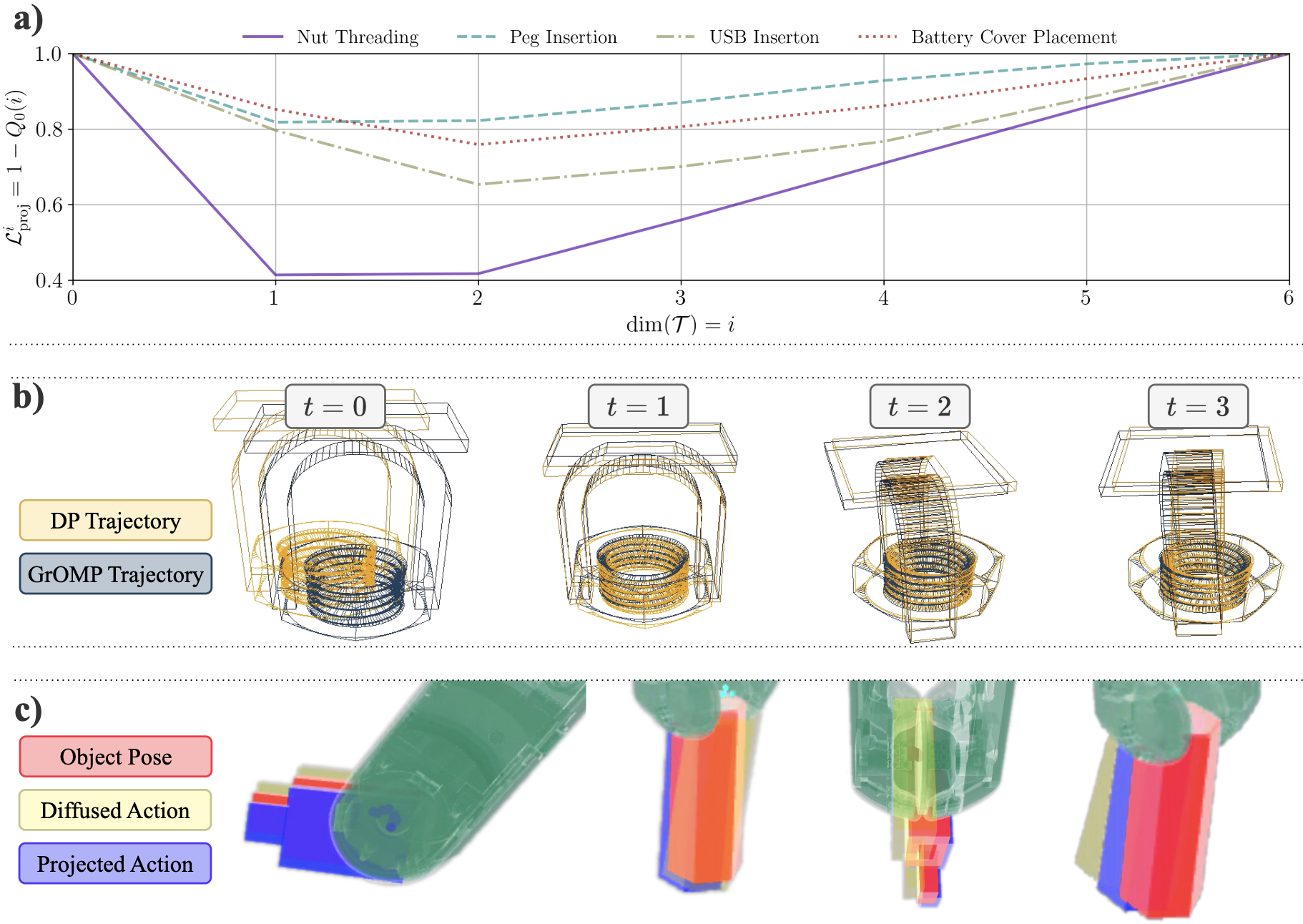}}
\caption{\textbf{a)} Projection loss priors (from Eq. \ref{eq:proj_loss}) derived from the dataset of each task tested in Section \ref{section:results}. \textbf{b)} Projection of the object and robot trajectories (as calculated in Section \ref{rollout}) in the expert dataset of nut threading along the manifold determined by Eq. \ref{eq:proj_param}, in this case $i^*_k=2$. \textbf{c)} Diffused vs. projected actions visualized for peg insertion and USB insertion alongside the current object pose observation. Here, the action is visualized as the last point in the action trajectory.}
\label{fig:projection}
\end{figure}

We then define the normalized twist $\vec{\xi}^*_n$, such that the rotational and translational components are of similar magnitudes:
\begin{equation}
    \vec{\xi}^*_n=
    \begin{bmatrix}
        \vec{\omega}/\text{max}(||\vec{\omega}||)\\
        \vec{v}/\text{max}(||\vec{v}||)
    \end{bmatrix}\label{eq:normalize_twist}
\end{equation}
We then perform principal component analysis (PCA) via singular value decomposition (SVD) on all normalized twists, represented as the twist matrix $\mat{\Xi}^*_{T\times6}=\begin{bmatrix}\vec{\xi}^{*,0}_n\ldots\vec{\xi}^{*,T-1}_n\end{bmatrix}^\top$, where $T$ is the number of data points. We assume the Euclidean mean of all twists is at the origin due to the use of geodesic mean in Eq. \ref{eq:convert}. This provides the PCA transform $\mat{P}_{6\times6}$:
\begin{align*}
    \mat{\Xi}^* &= \mat{U}\mat{\Sigma}\mat{P}^\top \text{ where } \mat{P} = \begin{bmatrix}
        \vec{p}_1\ldots\vec{p}_6
    \end{bmatrix}
\end{align*}
We can define a set of $\mat{\Xi}^i_P$ candidate projected twists by removing the principal degrees of freedom of the least significance, where $i=0,\ldots6$:
\begin{equation*}
    \mat{\Xi}_P^i = \mat{\Xi}^*\begin{bmatrix}
        \vec{p}_1\ldots\vec{p}_i&\mat{0}_{6\times(6-i)}
    \end{bmatrix}
\end{equation*}
which implies that $\mat{\Xi}_P^0 = \vec{0}_{T\times6}$ (a case of no projection). We then map the projected vector back to the original normalized tangent space defined by Eq. \ref{eq:normalize_twist}:
\begin{equation*}
    \mat{\Xi}^i = \mat{\Xi}_P^i\mat{P}^\top \text{ where } \mat{\Xi}^i=\begin{bmatrix}\vec{\xi}^{i,0}_n\ldots\vec{\xi}^{*i,T-1}_n\end{bmatrix}^\top
\end{equation*}
Note that all $\vec{\xi}^0_n = \vec{0}$, and all $\vec{\xi}^6_n = \vec{\xi}^*_n$. Less degrees of freedom ensures more predictable behavior of the robot, but we emphasize that some degrees of freedom are necessary to complete the task. For this reason, we define a loss $\mathcal{L}_{\text{proj}}^i$ (visualized in Fig \ref{fig:projection}) for each possible projection $i\in[0\ldots6]$:
\begin{equation}\label{eq:proj_loss}
    \mathcal{L}_{\text{proj}}^i = \frac{\sum_{t=0}^{T-1}\|\vec{\xi}^{*,t}_n-\vec{\xi}^{i,t}_n\|}{\sum_{t=0}^{T-1}\|\vec{\xi}^{*,t}_n\|} + \frac{i}{6}
\end{equation}
This loss trades off removing degrees of freedom vs accumulating errors. Intuitively, the more constraints the robot has, the less error accumulation at the cost of expressivity. This loss provides a prior for how to best select a projection for a given task; however, it could be sub-optimal due to the dataset or inference procedure. Thus, in the next section, we describe an interactive method to build upon this prior for a given task.

\subsection{7-Arm Bandit Adjustment}\label{interactive}
As stated in the previous section, Eq.~\ref{eq:proj_loss} provides a potentially suboptimal prior for how many principal components should describe the task manifold $\mathcal{T}$. Here, we describe an interactive formulation to infer the manifold, starting from the prior given by Eq.~\ref{eq:proj_loss}. To achieve this, we formulate the problem as a nonstationary $n$-arm bandit \cite{RLBook}. We first initialize the value $Q_k(i)$ of each projection as $Q_0(i)=1-\mathcal{L}_{\text{proj}}^i$. The reward of a selected projection $R_k(i^*_k)$ at each iteration $k$ is derived from $\alpha_k(i^*_k)$ successes and $\beta_k(i^*_k)$ failures after $K = \alpha_k(i^*_k)+\beta_k(i^*_k)$ task attempts:
\begin{equation*}
    R_k(i^*_k) = \frac{\alpha_k(i^*_k)}{\alpha_k(i^*_k)+\beta_k(i^*_k)}
\end{equation*}

Then, we update the value of the selected projection as follows, with step size $\gamma$:
\begin{equation}\label{eq:value_update}
    Q_{k+1}(i^*_k) = Q_{k}(i^*_k)+\gamma[R_k(i^*_k)-Q_k(i^*_k)]
\end{equation}
We use an $\epsilon$-greedy method \cite{RLBook} to obtain $i^*_k$ after each $K$ trials, selecting the projection with the highest value:
\begin{equation}
    i^*_k = \argmax_{i\in[0\ldots6]}Q_k(i)\label{eq:proj_param}
\end{equation}
We use Eq.~\ref{eq:proj_param} to select $i^*_k$, except when we make random selection for $i^*_k\in\{0\ldots6\}$ with a probability of $\epsilon$. We run this iteration over policy rollouts and continuously update our belief over the $Q$ estimates.

\subsection{Trajectory Rollout}\label{rollout}
In this section, we describe how to obtain $\text{proj}_\taskspace\affine_{so}$, which is necessary to calculate the robot poses $\affine_{st}$ for policy execution (as stated in Eq. \ref{eq:taskspaceconstraint}). In order to do this, we first select the task manifold described by dimensionality $i^*_k$ via the $n$-armed bandit approach described in the previous section. We then represent the object pose trajectories predicted by the learned policy $\affine_{so}=\affine_{st}^{\solpolicy}\affine_{to}$ as a normalized twist matrix $\hat{\mat{\Xi}}$, following the procedure from Sec.~\ref{selection}. Continuing this procedure, we project $\hat{\mat{\Xi}}$ to the reduced PGA space with $i_k^*$ columns of $\mat{P}$, then map back to the original tangent space with $\mat{P}^\top$. Finally, we convert the resulting twists back to $\se3$ using the exponential map:
\begin{align*}
    \begin{bmatrix}\vec{\xi}^{i^{*}_{k},0}_n\ldots\vec{\xi}^{i^{*}_{k},T-1}_n\end{bmatrix}^\top&=\hat{\mat{\Xi}}\begin{bmatrix}
        \vec{p}_1\ldots\vec{p}_{i^{*}_{k}}&\mat{0}_{6\times(6-{i^{*}_{k}})}
    \end{bmatrix}\mat{P}^\top\\\text{ where }\vec{\xi}^{i^{*}_{k}}&=
    \begin{bmatrix}
        \vec{\omega}^{i^{*}_{k}}\\
        \vec{v}^{i^{*}_{k}}
    \end{bmatrix}, \text{ and}\\
    \text{proj}_\taskspace\affine_{so}^{\solpolicy} &= \overline{\affine_{so}^{\exppolicy}}\exp(\hat{\vec{\xi}^{i^{*}_{k}}})\\ \text{ where }\hat{\vec{\xi}^{i^{*}_{k}}}&=\begin{bmatrix}
        \vec{\hat{\omega}}^{i^{*}_{k}}\text{max}(||\vec{\omega}||) & \vec{v}^{i^{*}_{k}}\text{max}(||\vec{v}||)\\
        \mat{0}_{1\times3} & 0
    \end{bmatrix}
\end{align*}
We show a grasped object's trajectory projected along this manifold in Fig. \ref{fig:projection}.

\section{Experimental Setup}\label{section:setup}
In this section, we describe the way in which we tested GrOMP using a tactile and proprioceptive representation of $\robotstate$, and a tactile-derived in-hand pose estimator such that $\affine_{to}$ is measurable in accordance with our assumptions stated in Section \ref{proj_filter}.
\subsection{Diffusion-Based Behavior Cloning} \label{subsection:DP}
Our policies are learned with Diffusion Policy (DP) \cite{chi2023}, using a cosine noise schedule \cite{Nic2021} with $\maxK$ forward steps. For fast rollout of policies, we use a DDIM \cite{son2022} with $\infK=\maxK/\TIscale$ reverse steps and $\eta$ parameter. To form a complete data point, an action $\actionind{0}$ is coupled with an observation feature $\obsfeature$ that is an embedding of multimodal feedback preceding time $\DI$. The noise prediction net $\noisenet$ is trained using a loss:
\begin{equation*}
   \loss_D=\sum\|\noise-\noisenet(\actionpred,\obsfeature,\NI)\|\label{eq:loss}
\end{equation*}
$\noisenet$ uses a 1D CNN U-Net architecture, as implemented in \cite{chi2023}. Upon rollout of the trained policy, $\horiz_e=4$ of the actions in $\actionind{0}$ are executed.

\subsection{Tactile Perception} \label{subsection:perception}
In this paper, vision-based tactile sensing from GelSlim 4.0 \cite{gelslim4} provides both one of the DP modalities (along with proprioception) and the means with which $\affine_{to}$ is measured. 
\subsubsection{Tactile Shear-Field} 
For DP, we found a shear-field based representation $\shearfield$ to be far more successful for our purposes than the raw GelSlim image representation. To derive the shear-field components $(\horizontal, \vertical)$, each a $13\times18$ matrix, from raw GelSlim RGB images $\image$, we used the \textbf{open-cv2} library. Our method uses a function $\optflowfunc$ which calculates the optical flow components from the undeformed frame $\image_0$ to the deformed frame $\image_\DI$: $(\horizontal, \vertical) = \optflowfunc(\image_0,\image_\DI)$. The full input to DP (including encoding) from both $L$ and $R$ fingers on our parallel jaw gripper is $\shearfield_{13\times 18\times4} = \{\horizontal^L,\vertical^L,\horizontal^R,\vertical^R\}$.
\subsubsection{Tactile In-Hand Pose Estimation}
\begin{figure}[h]
\centerline{\includegraphics[width=0.9\columnwidth]{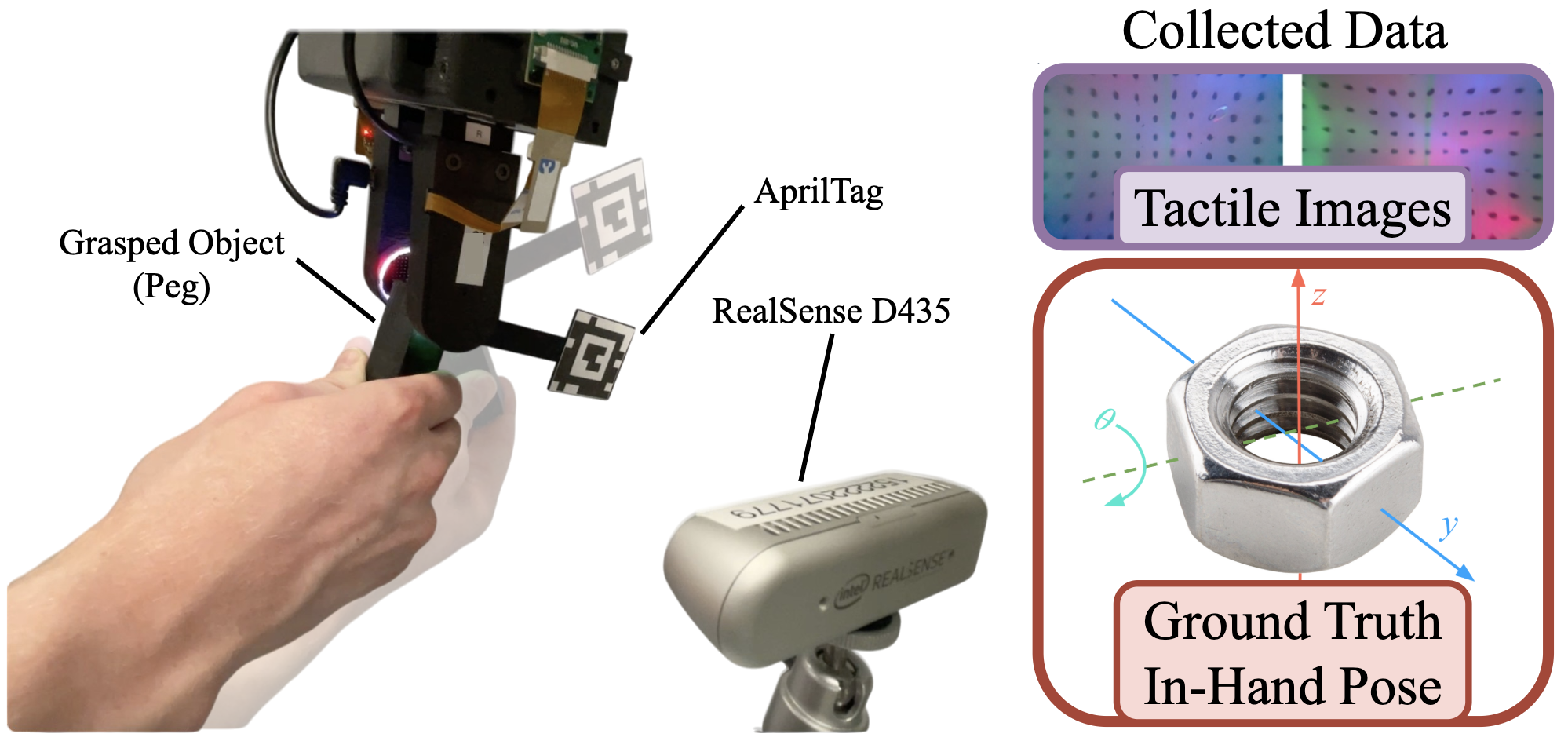}}
\caption{The tactile and ground-truth $\se{2}$ object pose data for in-hand pose estimation is collected while the object is grasped between the tactile sensors, and a human manually moves the object in the grasp. The grasp occasionally opens during this collection. Ground-truth object pose data comes from AprilTag registration using a RealSense D435 camera.}
\label{fig:ihp_dataset}
\end{figure}

To recover $\affine_{to}$ for GrOMP, we use an in-hand pose estimation module $\ihp$, consisting of a convolutional variational auto-encoder \cite{vae} and MLP, with layers as listed in Table \ref{tab:ihp_layers}. These are trained simultaneously on reconstruction of the raw RGB GelSlim image (downsampled 16x, 6-channel from two fingers), minimization of KL-divergence of the latent space, and regression to ground truth object poses. This is encompassed in the loss $\loss_{\text{ihp}}$:
\begin{align*}
   \loss_\text{ihp} &= w_\text{rec}\sum\|\{\image_R,\image_L\}-\{\image_R,\image_L\}_\text{rec}\|\\&+w_\text{kl}KL(N(\mu, \Sigma) \| N(0,1))\\&+\sum\|\proprio_{\text{ihp}}-\proprio^{\text{gt}}_{\text{ihp}}\|\label{eq:ihploss}
\end{align*}
Where the in-hand pose $\proprio_{\text{ihp}} = (y_{to},z_{to},\theta_{to})$, i.e. we restrict $\affine_{to} \in \se{2}$. We assume in-hand poses outside of this plane are minimal due to our parallel-jaw gripper. Additionally, the latent space $\mathbf{z_I}\sim N(\mu,\Sigma)$ is derived from the encoding: $(\mu,\Sigma) = \ihp_\text{Enc}\{\image_R,\image_L\}$, and is the input for the decoding $\{\image_R,\image_L\}_\text{rec} = \ihp_\text{Dec}(\mathbf{z_I})$ and the prediction $\proprio_{\text{ihp}}=\ihp_\text{MLP}(\mathbf{z_I})$. The dataset for training this in-hand pose estimation module is collected via apriltag registration, frequent regrasping, and manual object manipulation, as shown in Fig. \ref{fig:ihp_dataset}.

\subsection{Observation Encoding}\label{subsection:obs_encoding3}
Observations to be encoded are normalized linearly to be in the domain $(-1,1)$ to work with the diffusion U-Net. The shear field $\shearfield$ is normalized according to the values found in the dataset. Each component channel $\horizontal^L,\vertical^L,\horizontal^R,\vertical^R$ is normalized to $(-1,1)$. The translational values within the cartesian proprioception $\affine_{st}$ are also normalized based on the values in the dataset, with each component normalized to $(-1,1)$. We convert the rotation matrix within $\affine_{st}$ to the 6D representation from \cite{zhou2019} which remains un-normalized through the entire process, as suggested for Diffusion Policy \cite{chi2023}. Thus, the entire representation for proprioception is the translation-normalized 9D vector $\proprio$.

We use an observation horizon of just $\obsperiod=1$. We use a convolutional neural network (CNN), $\vfencoder$ to encode shear-field observations, the layers of which are described in Table \ref{tab:cnn_layers}. $\proprio$ is directly concatenated into $\obsfeature$:
\begin{align*}
    \obsfeature = \vfencoder(\shearfield_\DI)\oplus \proprio_\DI
\end{align*}

\begin{table}[t]
\begin{center}
\caption{$\ihp$ layers}
\begin{tabularx}{\columnwidth}{*{2}{l} *{1}{Y}}
\toprule
Segment  & Layer & Description\\
\midrule
\multirow{6}{*}{\specialcell{Encoder\\$\ihp_{\text{Enc}}$}}& Conv2D & In: $6$, Out: $12$, $3\times3$, MaxPool($2$)\\
& Conv2D & In: $12$, Out: $32$, $3\times3$, MaxPool($2$)\\
& Flatten & From size $H\times W\times 32$ to size $h$\\
& Linear & In: $h$, Out: $256$\\
& 2$\times$Linear & In: $256$, Out: $\mu:256,\Sigma:256$\\
& Sample $\sim N(\mu,\Sigma)$ & In: $\mu:256,\Sigma:256$, Out: $256$\\
\midrule
\multirow{2}{*}{\specialcell{Predictor\\$\ihp_{\text{MLP}}$}} & Linear & In: $256$, Out: $84$\\
& Linear & In: $84$, Out: $3$\\
\midrule
\multirow{6}{*}{\specialcell{Decoder\\$\ihp_{\text{Dec}}$}} 
& Linear & In: $256$, Out: $256$\\
& Linear & In: $256$, Out: $h$\\
& Reshape & From size $h$ to size $H\times W\times 32$\\
& ConvTranspose2D & In: $32$, Out: $12$, $4\times4$, MaxPool($2$)\\
& ConvTranspose2D & In: $12$, Out: $6$, $4\times4$, MaxPool($2$)\\
\bottomrule
\end{tabularx}
\label{tab:ihp_layers}
\end{center}
\end{table}

\begin{table}[b]
\begin{center}
\caption{$\vfencoder$ layers}
\begin{tabularx}{\columnwidth}{l *{1}{Y}}
\toprule
Layer & Description\\
\midrule
Conv2D & In: $4$, Out: $16$, $4\times4$, MaxPool($2$)\\
Conv2D & In: $16$, Out: $64$, $4\times4$, MaxPool($2$)\\
Flatten & To Size $h$\\
Linear & In: $h$, Out: $128$\\
\bottomrule
\end{tabularx}
\label{tab:cnn_layers}
\end{center}
\end{table}

\subsection{Task Demonstrations}\label{subsection:demos3}
We collect a tactile and proprioceptive dataset for each task the robot is to perform; both are important to our implementation of both GrOMP and IL.

\subsubsection{Demonstration Collection}\label{subsection:demo_collection3}
\begin{figure}[h]
\centerline{\includegraphics[width=\columnwidth]{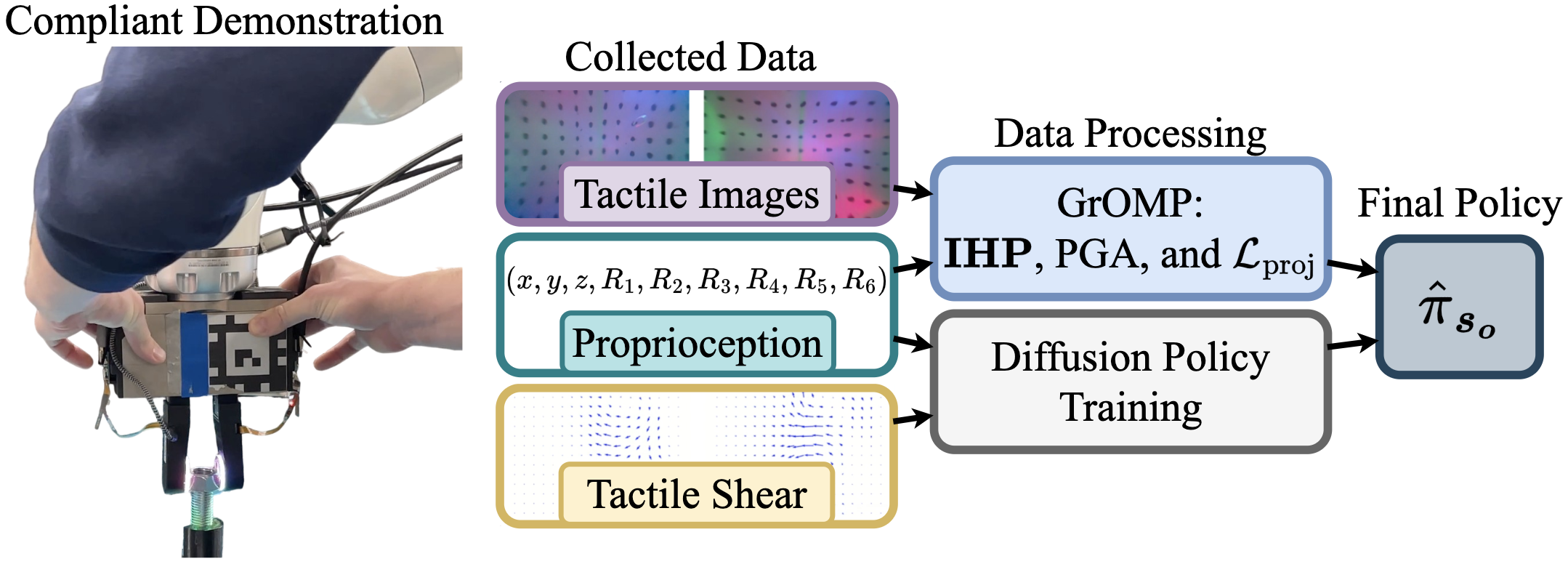}}
\caption{View of manual demonstration as described in Section \ref{subsection:demo_collection3}, and the pipeline of the tactile and proprioceptive data toward the generation of the combined policy $\combined$.}
\label{fig:demonstration}
\end{figure}

We collect kinesthetic demonstrations since this methodology is particularly well suited to both precise assembly tasks and tactile sensing. To collect kinesthetic demonstrations, the robot is first placed in a compliant state using impedance control. Then, the human expert manually manipulates the end effector of the robot towards task completion, as shown in Fig. \ref{fig:demonstration}.

\subsubsection{Demonstration Processing}\label{subsection:demo_processing3}
All demonstration episodes for behavior cloning a single task are sampled such that each episode contributes $\horiz_E=64$ action/observation pairs, where $\actionind{0}$ consists of $\horiz_a=8$ cartesian robot poses $\affine_{st}$ (converted to 9D pose) within the $\se3$ robot workspace. Using position (as opposed to velocity) as an action space in this way is consistent with \cite{chi2023} and lends itself well to GrOMP. Actions are normalized to $(-1,1)$ just as observations. We augment the dataset 8 times by adding noise $\sim N(0,0.1)$ to the normalized tactile shear-field. We use a train-validation split ratio of 80:20. For every 10 episodes, we then have 4096 action-observation pairs to train, and 1024 to validate. The weights of $\noisenet$ that produce the lowest $\mathcal{L}_D$ during training are selected for testing.

\section{Experiments and Results}\label{section:results}
\begin{figure}[t]
\centerline{\includegraphics[width=\columnwidth]{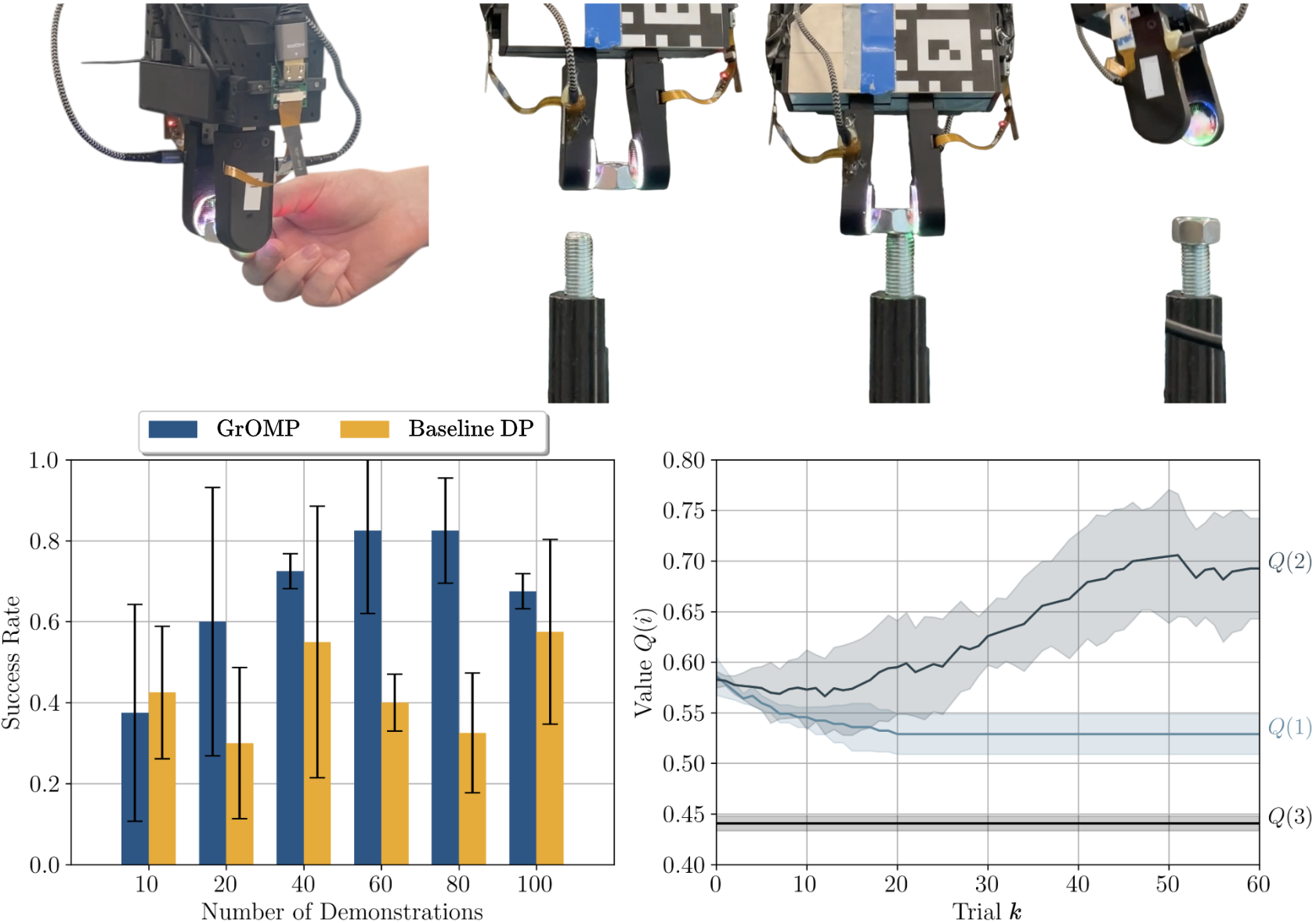}}
\caption{Nut threading results. \textbf{Top:} Handoff, initialization, policy, and success snapshots. \textbf{Left:} GrOMP vs DP performance results as demonstrations are added to training. \textbf{Right:} Change of highest values in $Q(i)$ from Section \ref{interactive} over the 60-trial horizon, averaged over the 4 runs of GrOMP. Filled area surrounding curves represents $\sigma$ (the initial projection loss in Eq. \ref{eq:proj_loss} sometimes yielded $i_k^*=1$ but $i_k^*=2$ was the eventual result in all 4 runs of USB insertion).}
\label{fig:nut_threading}
\end{figure}

\begin{figure}[b]
\centerline{\includegraphics[width=\columnwidth]{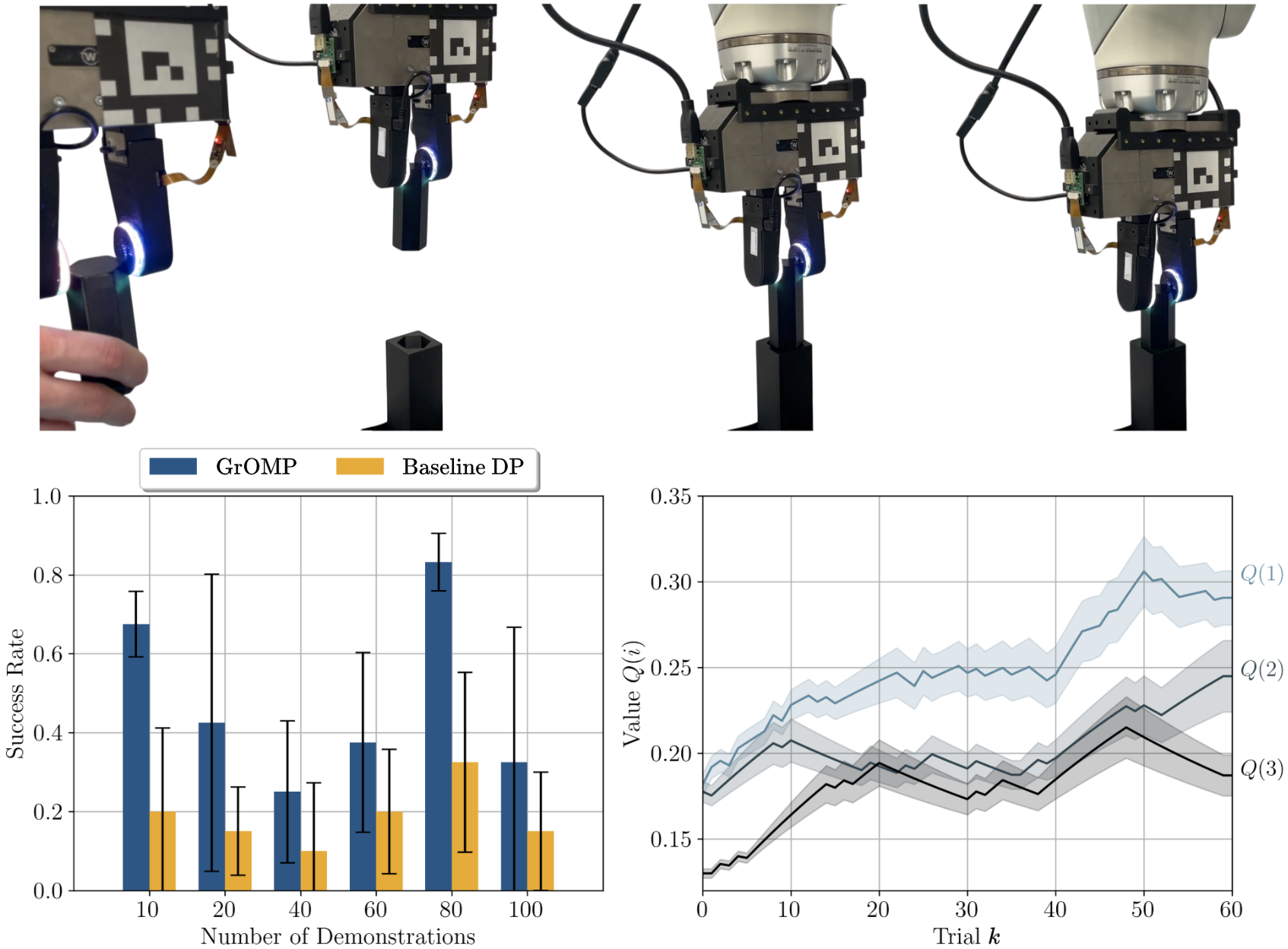}}
\caption{Peg insertion results. \textbf{Top:} Handoff, initialization, policy, and success snapshots. \textbf{Left:} GrOMP vs DP performance results as demonstrations are added to training. \textbf{Right:} Change of highest values in $Q(i)$ from Section \ref{interactive} over the 60-trial horizon, averaged over the 4 runs of GrOMP. Filled area surrounding curves represents $\sigma/8$ (high variance in the result of Eq. \ref{eq:proj_param} occurred between the 4 runs of peg insertion).}
\label{fig:peg_insertion}
\end{figure}

We tested GrOMP as described in Section \ref{method} with DP as described in Section \ref{section:setup}, against pure vanilla DP as a baseline, with four precise assembly tasks: nut threading, peg insertion, USB insertion, and battery cover placement. We treated this as an interactive imitation learning trial where demonstration episodes are added every 10 trials to train new policies. For the DP baseline, we just tested the effect of adding episodes, beginning with 10, then 20, 40, 60, 80, and finally 100. For GrOMP we tested the same, with the effect of the projection to $\taskspace$ (Eq. \ref{eq:taskspaceconstraint}). The initial value $Q_0(i)$ was determined by the first 10 episodes only. We set the value update in Eq. \ref{eq:value_update} to occur every $K=1$ trial, with $\gamma=0.025$. The $\epsilon$-greedy method for selecting the projection dimensionality $i_k^*$ is performed with $\epsilon=0.1$. Each experiment---a full set of 6 different amounts of training episodes, tested 10 times each---is replicated 4 times. For each of these 4 replications, episodes are introduced in a new random order. As a result, each task is tested with \textbf{240 runs for GrOMP and 240 runs for the DP baseline}. Each run is given a maximum policy horizon: $T_E=64$ action prediction steps to complete the task. Snapshots of the four tasks we test are found in Figs. \ref{fig:nut_threading}–\ref{fig:battery_cover}.

\subsection{Nut Threading}
For nut threading, the object to be assembled is an M20 nut which must be mated with an M20 bolt fixed vertically to the environment. The task is considered successful if the nut cannot be removed from the bolt via a vertical lift (i.e. it must be twisted off). Fig.~\ref{fig:nut_threading} shows nut threading results.

\subsection{Peg Insertion}
For peg insertion, the object to be assembled is a 25 mm wide hexagonal prism peg which must be inserted into a hexagonal hole with 0.25 mm radial clearance, fixed vertically to the environment. The task is considered successful if the peg falls to the bottom of the hole when released by the gripper. Results for peg insertion can be seen in Fig. \ref{fig:peg_insertion}.

\begin{figure}[t]
\centerline{\includegraphics[width=\columnwidth]{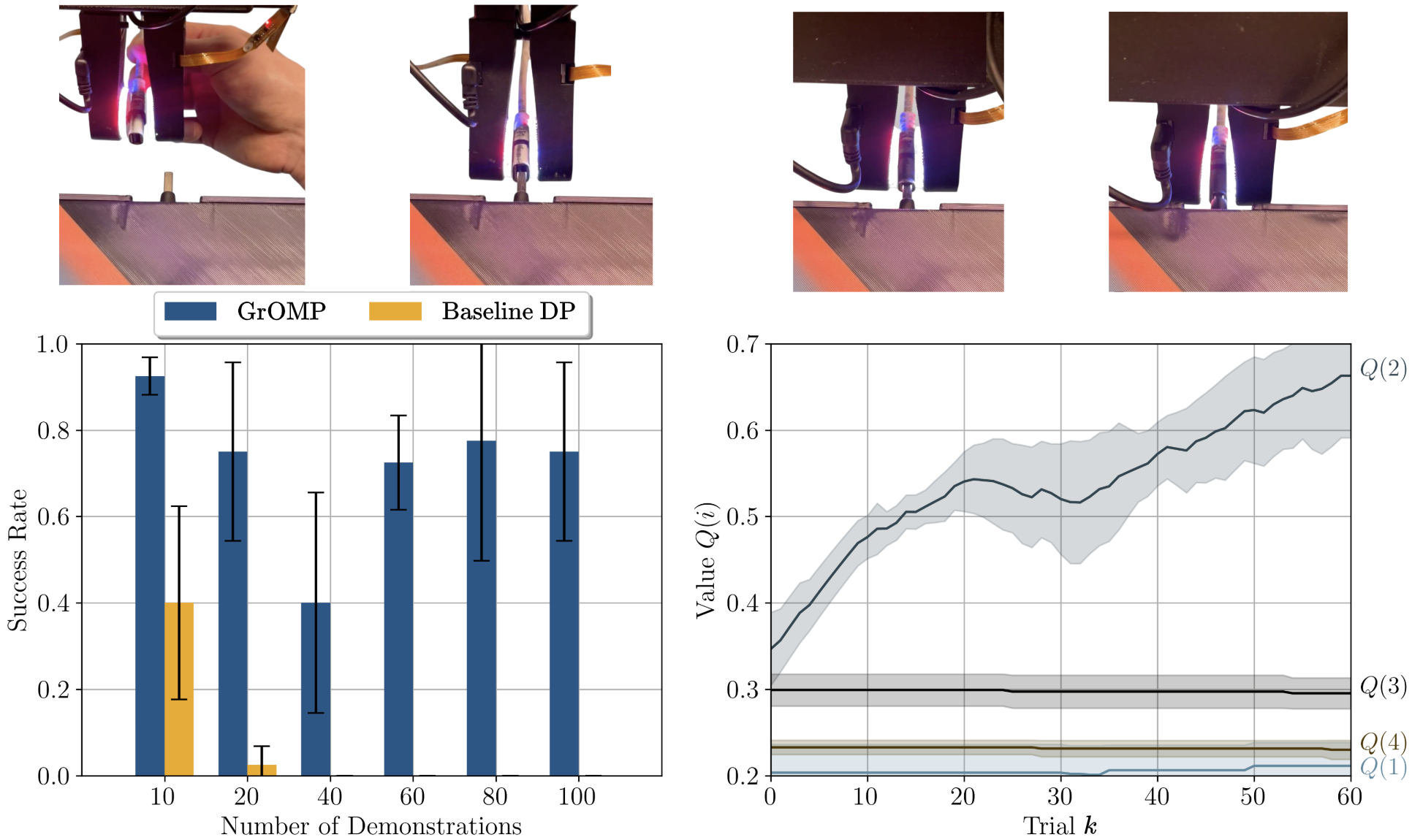}}
\caption{USB insertion results. \textbf{Top:} Handoff, initialization, policy, and success snapshots. \textbf{Left:} GrOMP vs DP performance results as demonstrations are added to training. \textbf{Right:} Change of highest values in $Q(i)$ from Section \ref{interactive} over the 60-trial horizon, averaged over the 4 runs of GrOMP. Filled area surrounding curves represents $\sigma$ (Eq. \ref{eq:proj_param} 
always yielded $i_k^*=2$ in all 4 runs of USB insertion).}
\label{fig:usb_insertion}
\end{figure}

\subsection{USB Insertion}
For USB insertion, the object to be assembled is the female end of a USB-A extension cable,  which must be inserted into a male USB connector fixed horizontally to the environment. The task is considered successful if a horizontal movement primitive can complete the insertion after the policy horizon has completed. Results for USB insertion can be seen in Fig. \ref{fig:usb_insertion}.

\begin{figure}[b]
\centerline{\includegraphics[width=\columnwidth]{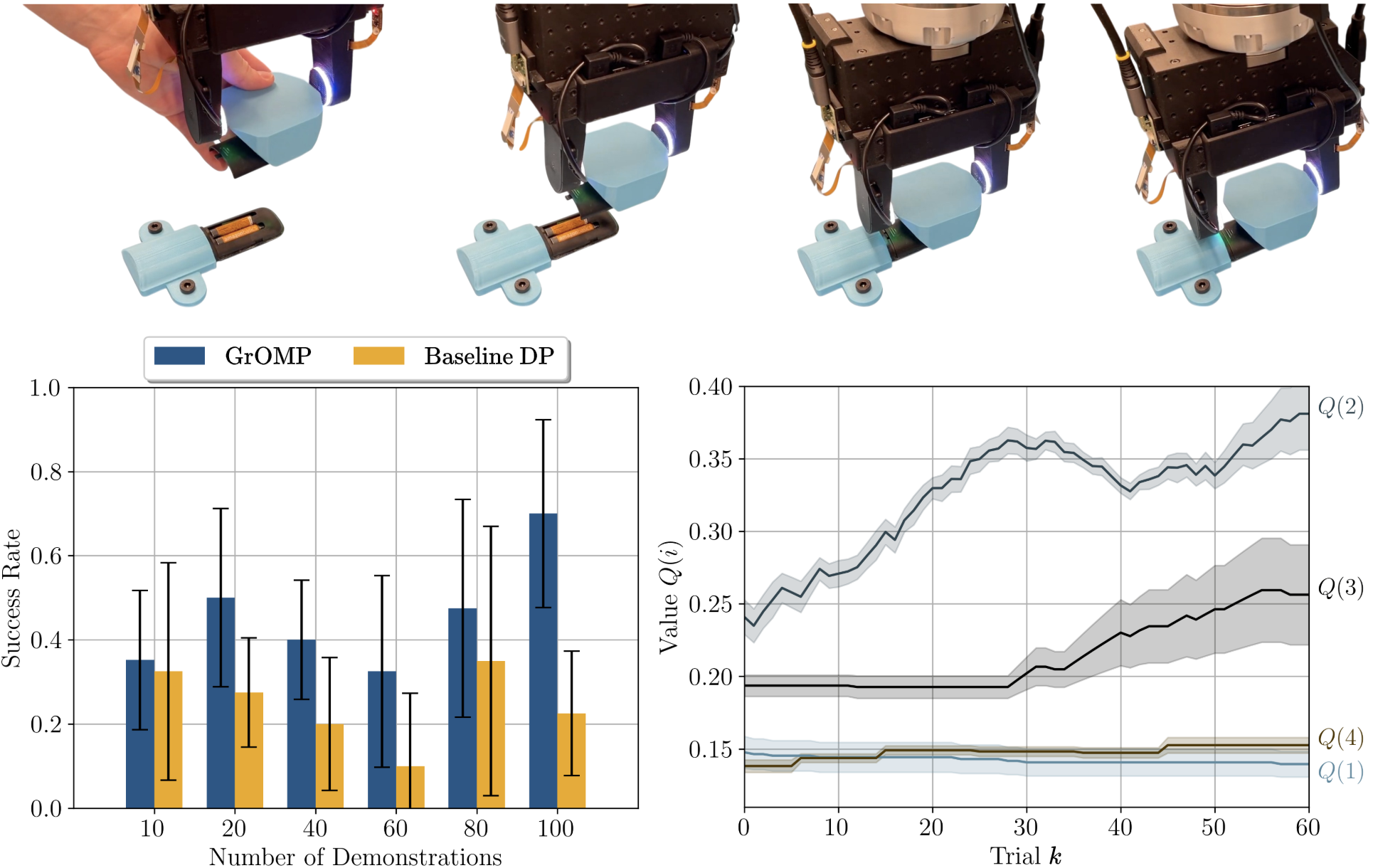}}
\caption{Battery cover placement results. \textbf{Top:} Handoff, initialization, policy, and success snapshots. \textbf{Left:} GrOMP vs DP performance results as demonstrations are added to training. \textbf{Right:} Change of highest values in $Q(i)$ from Section \ref{interactive} over the 60-trial horizon, averaged over the 4 runs of GrOMP. Filled area surrounding curves represents $\sigma$ (Eq. \ref{eq:proj_param} 
yielded $i_k^*=2$ most commonly, with $i_k^*=3$ being selected near the end of 1 of 4 runs of battery cover placement).}
\label{fig:battery_cover}
\end{figure}

\subsection{Battery Cover Placement}
For battery cover placement, the object to be assembled is a battery cover for a Roku remote control, which must be mated with the remote control body fixed horizontally to the environment. The task is considered successful if a horizontal movement primitive can complete the mating after the policy horizon has completed. Results for battery cover placement can be seen in Fig. \ref{fig:battery_cover}.

\section{Discussion}
Our results suggest that GrOMP improves task performance over vanilla Diffusion Policy (DP). One key effect of GrOMP is the tendency to avoid unrecoverable states in contrast to vanilla IL, especially when the object shifts in the grasp. For example, nut threading with DP alone may exhibit behavior where the nut contacts the bolt such that it tilts heavily within the grasp. GrOMP allows the robot to position itself such that the nut remains vertical, even though this robot configuration does not occur in DP's expert dataset. 

GrOMP's behavior is beneficially reactive to disturbances. This same kind of tilting can happen in peg insertion, USB insertion, and battery cover placement as well. A secondary recovery strategy requires obstacle avoidance. For instance, our peg insertion task is tested with a hole that does not have an excess of surface surrounding its mouth, instead there is empty space. If the peg is erroneously plunged into this empty space rather than the hole, a re-lift of the peg is required to recover, but this behavior does not appear in the expert dataset, nor is it likely to occur with GrOMP's projection. GrOMP is more likely to prevent this kind of erroneous behavior to begin with, leading to our more successful peg insertion results seen in Fig. \ref{fig:peg_insertion}.

\subsection{Anomalous Baseline Performance}
We note that our results do not always show a positive relationship between the number of demonstrations and task success rates which contradicts conventional wisdom. We particularly did not expect the seemingly negative trend between these variables for our USB insertion results, where a policy trained on 10 demonstrations can achieve around a 40\% success rate while policies trained on more data all but fail completely.

This type of performance degradation has presented in experimental imitation learning literature before. Specifically, in proposing DAgger \cite{ross2011_dagger}, Ross et al. compared the interactive behavior cloning algorithm with a supervised learning baseline (Diffusion Policy is our supervised learning baseline). Their results showed first no improvement with an increasing number of demonstrations, and then a drop in success rate when the policy is presented with further training examples. Their explanation for this phenomenon stated that similar demonstrations being introduced to training cannot be expected to improve performance, but did not attempt to explain the drop in performance.

Speaking specifically to the task of USB insertion, one paper by George et al. \cite{geo2024} investigated behavior cloning of this task against multiple sensing modalities (visual, tactile, and both) and multiple behavior cloning methods (Diffusion Policy and Action  Chunking Transformers) \cite{geo2024}. They fixed the number of task demonstrations at 100, providing us with a close comparison to one combination of factors from our own experiments. Matching our results, they showed that a tactile-only diffusion policy trained on 100 demonstrations achieved no success. Perhaps 10 demonstrations under their implementation would produce some success.

The reason for this paradoxical outcome is debatable, though we hypothesize some form of overfitting is occurring, possibly due to suboptimal, repetitive, or noisy demonstrations. There are also several hyperparameter and architecture choices when it comes to Diffusion Policy. We do note that testing over this set is prohibitively expensive given the need for real-world rollouts. Our results show that GrOMP successfully improves upon Diffusion Policy for these tasks, independent of the possible existence of suboptimal demonstrations or design choices.

\subsection{Limitations}
Future improvements may be inspired by GrOMP's limitations. 
GrOMP currently does not learn a temporal manifold, meaning projecting to $\taskspace$ will not necessarily progress the task temporally. DP retains full responsibility for task completion, while GrOMP attempts to prevent accumulating errors. GrOMP also requires a reliable in-hand pose estimator. Any improvements in the pose estimation method will serve to improve GrOMP. For instance, we demonstrated a predictable $\se{2}$ estimator given our assumptions (Sec.~\ref{subsection:perception}). GrOMP can be improved if $\ihp$ were expanded to $\se{3}$ as some works have attempted ~\cite{t3}. We note that tactile may not be available for all systems. A secondary solution would be to use vision techniques to estimate poses; however, these techniques are known to struggle with occlusion. 

Our hope is that the inspiration behind GrOMP leads to the consideration of more geometry-based methods on top of the benefits that imitation learning already provides. Along this research path, the data efficiency and precision required by manipulation in fixtureless industrial assembly can be fully realized.


\bibliographystyle{ieeetr}
\bibliography{references.bib}

\end{document}